\documentclass[runningheads]{llncs}
\usepackage[T1]{fontenc}
\usepackage{graphicx}
\usepackage{algorithm}
\usepackage{algorithmic}
\usepackage{latexsym}
\usepackage{amssymb}
\usepackage{amsmath}
\usepackage{booktabs}
\usepackage{enumitem}
\usepackage{graphicx}
\usepackage{color}
\usepackage{amsfonts}
\usepackage{array}
\usepackage{textcomp}
\usepackage{stfloats}
\usepackage{url}
\usepackage{verbatim}
\usepackage{epsfig}
\usepackage{multirow}
\usepackage{bbding}
\usepackage[marginal]{footmisc}

\begin{document}
\title{Granular-ball Representation Learning for Deep CNN on Learning with Label Noise}
%
%
\author{Dawei Dai\inst{1(}\Envelope\inst{)}\orcidID{0000-0002-8431-4431} \and
Hao Zhu\inst{1}\orcidID{0000-0002-4655-7336} \and
Shuyin Xia\inst{1}\orcidID{0000-0001-5993-9563} \and
Guoyin Wang\inst{1}\orcidID{0000-0002-8521-5232}}
\authorrunning{F. Author et al.}
%
\institute{Chongqing Key Laboratory of Computational Intelligence, Key Laboratory of Big Data Intelligent Computing, Key Laboratory of Cyberspace Big Data Intelligent Security, Ministry of Education, Chongqing University of Posts and Telecommunications, Chongqing, 400065, China}
\maketitle              

\footnote{First Author and Second Author contribute equally to this work.\\}
%
\begin{abstract}
In actual scenarios, whether manually or automatically annotated, label noise is inevitably generated in the training data, which can affect the effectiveness of deep CNN models. The popular solutions require data cleaning or designing additional optimizations to punish the data with mislabeled data, thereby enhancing the robustness of models. However, these methods come at the cost of weakening or even losing some data during the training process. As we know, content is the inherent attribute of an image that does not change with changes in annotations. In this study, we propose a general granular-ball computing (GBC) module that can be embedded into a CNN model, where the classifier finally predicts the label of granular-ball ($gb$) samples instead of each individual samples. Specifically, considering the classification task: (1) in forward process, we split the input samples as $gb$ samples at feature-level, each of which can correspond to multiple samples with varying numbers and share one single label; (2) during the backpropagation process, we modify the gradient allocation strategy of the GBC module to enable it to propagate normally; and (3) we develop an experience replay policy to ensure the stability of the training process. Experiments demonstrate that the proposed method can improve the robustness of CNN models with no additional data or optimization.
\end{abstract}
%
%
\section{Introduction}

In recent years, deep CNN models have achieved great success in many fields owing to their powerful feature representation and learning abilities \cite{10028728}. However, their usefulness is usually dependent on high-quality annotated data. Typically, two common data-annotation methods can be used, that is manual and automatic model annotation \cite{benato2021semi}. Both of them are inevitably bound to produce a certain proportion of wrong annotation data (label noise) owing to myriad constraints, including the professional domain knowledge of the annotation personnel, data quality, malicious data poisoning, and the performance of the annotation model. Excessive mislabeled data (label noise) can cause changes or even confusion in the distribution of the training data, leading to diminished performance or even specific bias discrimination in related tasks \cite{zhang2021understanding}. Consequently, constructing a CNN model robust to label noise is of practical significance. \par

Currently, two main solutions are employed for label noise, that is noise containment and noise filtering. The former refers to reducing the impact of noisy labels by designing the additional optimizations to punish samples with wrong labels. Noise filtering involves clearing or correcting noise samples before returning them to the model for training, which has problems of its own. For example, (1) when the proportion of label noise is high, clearing all noisy samples greatly reduces the size of the training dataset, which can lead to insufficient or imbalanced training samples, and (2) this type of method usually targets low-dimensional samples and can have difficulty handling high-dimensional samples such as images. Our aim is to develop a general module that can be embedded in the CNN models to improve their robustness. \par
\begin{figure}[t]
	\centerline{\includegraphics[width=0.60 \textwidth]{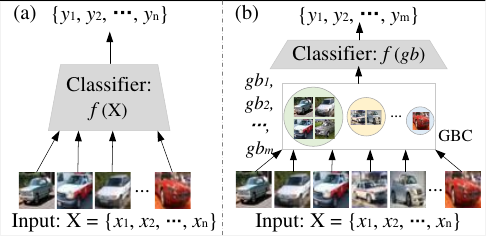}}
	\caption{Illustration of the proposed GBC module. (a) Traditional mode of mapping each individual sample to one label; (b) GBC mode of mapping each granular-ball to one label. 
	}
	\label{fig0}
\end{figure}

The content and feature space of samples are inherent attributes, whereas labels are generated by human induction and definition. Therefore, samples can often be mislabeled, while the content of the sample or its feature space does not change with changes in the labeling. As we know, traditional classifiers learn the mapping of each individual sample that the single and finest granularity to its label, thus, the label noise has a major impact on the models. When the classifier learns the mapping of the cluster samples (granular-ball) at feature level based on content similarity to one label, that is, when multiple samples share one label (See Fig. \ref{fig0}), it substantially reduces the impact of single-sample label noise on the model; if the label noise ratio of granular-ball samples is much lower than that of individual samples, it substantially reduces the impact of single-sample label noise on the models.

In this study, we develop a novel GBC module for the CNN models to learn with noisy labels, aiming to improve robustness of CNN with no additional requirements to the original model. GBC, first proposed by Xia. et al. \cite{xia2020fast}, is considered to be an effective method for describing a multi-granularity knowledge space. However, current GBC methods focus only on areas such as statistical machine learning and rough sets \cite{xia2022random,xia2022gbrs,xia2024efficient}. We extend the GBC to split the hidden feature vectors of the input batch samples into granularity-ball samples for deep CNN models. Specifically, our GBC module splits the input into granularity-ball ($gb$) samples at feature-level, each of $gb$ sample contains an unequal number of individual samples with sharing one label. The specific label shared by the most of individual samples in one $gb$ sample can be assigned as this $gb$ sample's label, and the classifier finally learns the mapping of each $gb$ sample to its label. Experiments show that the proportion of $gb$ sample's label noise is much lower than that of the individual sample, and that our proposed method can improve the robustness of original CNN models for image classification tasks. Our contributions can be summarized as:

(1) We develop a general GBC module that can be embeded into a CNN model to learn the multi-granularity representation for the classifier, where the traditional mode of learning the mapping from each individual sample to one label is transformed into a multi-granularity (MG) mapping of each $gb$ sample to its label.

(2) Our proposed GBC module can be embedded into a CNN model with no additional design and enhance the robustness of the original model on learning with label noise. When the GBC module is applied to a contrastive learning framework, it achieves the state-of-the-art results.

\section{Related Work}
\subsection{Noise Filtering}
A direct approach to deal with label noise is to design a specific method to remove mislabeled data. Han et al. \cite{han2018co} proposed a co-learning noise memory method, in which two networks with different learning capabilities were designed to perform collaborative learning on small batches of data to filter noise label samples. Guo et al. \cite{guo2018curriculumnet} developed principled learning strategies to achieve the goal of effectively dealing with a large number of noisy data label imbalances. Jiang et al. \cite{jiang2018mentornet} proposed learning other types of neural networks, called MentorNet, to supervise the training of basic deep networks (i.e., StudentNet), during which MentorNet could provide StudentNet with a course (sample weight scheme) to focus on samples with potentially correct labels. Jie et al. \cite{huang2019o2u} proposed making the learning rate change periodically—the model swinging between overfitting and underfitting, resulting in the loss of samples with noise labels changing considerably—to detect noise label samples. Jindal et al. \cite{jindal2019effective} introduced a nonlinear processing layer to model the data with incorrect labels, thereby preventing the model from overfitting noise. Yao et al. \cite{yao2024searching} considered that co-learning could not accurately express the true learning status of a network by manually setting the forgetting rate, and proposed to adaptively obtain the forgetting rate and enhance its autonomy. Liang et al. \cite{liang2020bond} proposed a two-stage training algorithm, that is, in the first stage, a pre-trained language model was adapted to named entity recognition (NER) tasks; in the second stage, remote label removal and self-training were used to enhance the robustness of the model. Meng et al. \cite{meng2021distantly} proposed a noise-robust learning scheme comprising a new loss function and noise label deletion process, training the model to label the data.  Garg et al. \cite{garg2021towards} proposed a two-component beta mixture model, assigning probability scores with clean or noisy labels to each sample before training the classifier and noise model using denoising losses. Zhang et al. \cite{zhang2021improving} proposed self-cooperative noise reduction learning, which trained a teacher-student network, with each network using reliable labels through self-denoising, and explored unreliable annotations through collaborative denoising. Li et al. \cite{Li2024FedDivCN} proposed a global noise filter called Federated Noise Filter(FedDiv) for effectively identifying samples with noisy labels.

\subsection{Noise Containment}
These methods attempt to design specialized optimization goals to construct robustness models. Manwani et al. \cite{manwani2013noise} verified that risk minimization using the 0–1 loss function had noise tolerance characteristics and the square error loss only tolerated uniform noise. Sukhbaatar et al. \cite{sukhbaatar2014training} introduced an additional noise layer in a neural network that adjusted the output to match the distribution of noise labels so that the probability transfer matrix continuously tended toward the true probability transfer matrix during the training process. Azadi et al. \cite{Azadi2015AuxiliaryIR} proposed an auxiliary image regularization technique, encouraging the model to select reliable images to improve the learning process. Jindal et al. \cite{jindal2016learning} augmented a standard deep network using a SoftMax layer that model the label-noise statistics before training the deep network. Zhuang ed al. \cite{zhuang2017attend} proposed an end-to-end weakly supervised deep-learning framework which was robust to label noise in web images. Li et al. \cite{li2017learning} proposed a unified distillation framework to use "edge" information to "hedge" the risk of learning from noisy labels. Patrini et al. \cite{patrini2017forward} proposed a forward correction method that does not depend on the application domain and network architecture, but only needs to know the probability of each class being polluted into another class. Zhang et al. \cite{zhang2018generalized} proposed a robust generalized cross-entropy (GCE) loss which combined the fast convergence speed of cross-entropy and the robustness advantages of the mean absolute error. Wang et al. \cite{wang2019symmetric} proposed a symmetric cross-entropy learning method that symmetrically enhances the CE using reverse cross-entropy corresponding to robust noise. Jun Shu et al. \cite{NEURIPS2019_e58cc5ca} proposed a meta-learning method to train a reliable network with a set of clean and small data to guide the subsequent training of noisy data, so as to alleviate the adverse effects of label noise or long-tail data on model training. Harutyunyan et al. \cite{pmlr-v119-harutyunyan20a} proposed a method to control the label noise information in the weights of neural networks, which reduced the label memorization problem. Ma et al. \cite{ma2020normalized} proposed to combine two mutually reinforcing robust loss functions to mitigate the underfitting problem and improve the learning performance. Chen et al. \cite{Chen2021NoiseAN} proposed a stochastic label noise (SLN) to help models avoid falling into "sharp minima" and "overconfidence" situations. Li et al. \cite{9879400} proposed a contrastive regularization function to learn robust contrastive representations over noisy data. Zhang et al. \cite{Zhang_2023_ICCV} proposed a representation calibration method, RCAL, which improves the robustness of the representation by recovering the multivariate Gaussian distribution.

\subsection{Granluar Computing}
Chen\cite{chen1982topological} pointed out that the brain gives priority to recognizing a "wide range" of contour information in image recognition, and human cognition has the characteristics of "global precedence". This differs from major existing artificial intelligence algorithms, which use the most fine-grained points as inputs. Granular computing can be used to partition data distribution and knowledge space. Wang \cite{wang2017dgcc} introduced a large-scale cognitive rule into granular computing and proposed multigranular cognitive computing. Xia and Wang \cite{xia2020fast} proposed hyperspheres of different sizes to represent "grains" and proposed GBC, in which a large $gb$ represented coarse granularity, while a small $gb$ represented fine-granularity. Xia et al. \cite{xia2022gbsvm} proposed the granular-ball support vector machine (GBSVM) method, in which $gb$ samples replaced the original finest-grained sample; this method exhibited better efficiency and robustness than the traditional classifier. GBC has also been applied in many other fields to improve model generalizability or efficiency, such as rough sets \cite{xia2022gbrs}, sampling \cite{xia2022random}, fuzzy sets \cite{xia2024efficient}. In this study, we develop an extended GBC framework to construct robust deep CNN models for learning with label noise.

\section{METHODOLOGY}
\subsection{Motivation}
\begin{figure*}[htbp]
	\centering
	\setlength{\belowcaptionskip}{0cm}
	\includegraphics[width=0.98\textwidth]{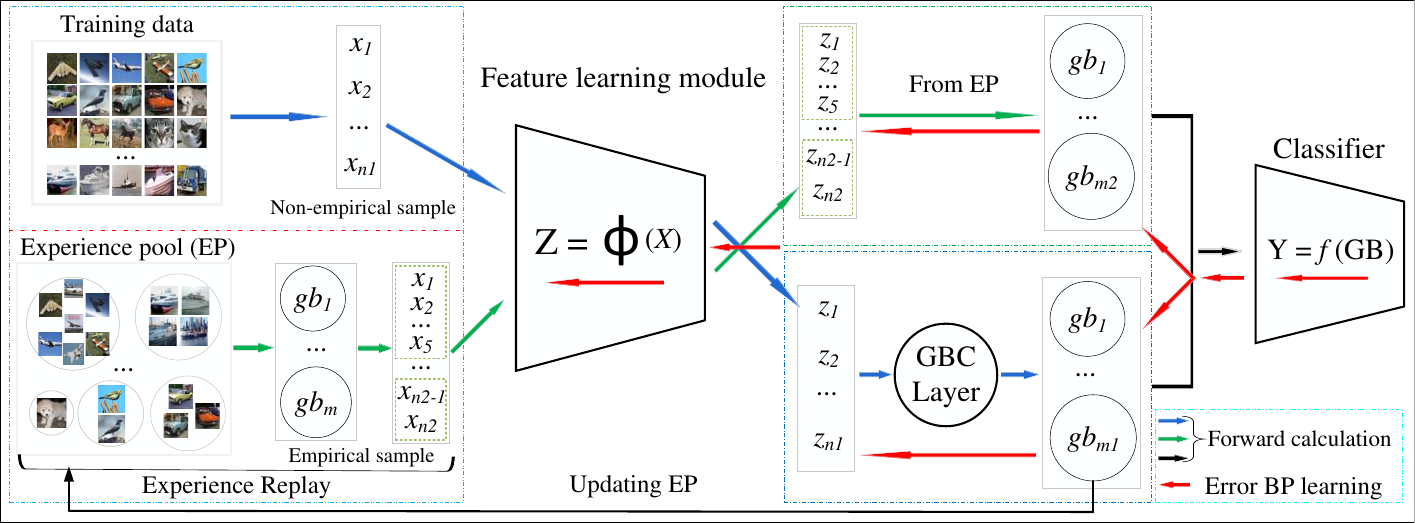}
	\caption{Overview of our proposed method.} 
	\label{fig3}
\end{figure*}

At present, the learning process of all deep CNN models attempts to map each individual sample in the training dataset to its label, that is, a single-granularity information processing mode. Therefore, containing a certain proportion of labeled noise in the training dataset can affect the usefulness of neural models. The popular solutions enhance the robustness of models at the cost of weakening or even losing the mislabeled data. In this study, we propose a GBC module that can be embeded in the CNN models, it splits the feature vectors of the input into multi-granularity grains ($gb$ samples). Consequently, the final classifier learns the mapping of each $gb$ sample to its label (Fig. \ref{fig3}). Intuitively, the proportion of $gb$ samples with incorrect labels that generated based on content similarity is unlikely to exceed or may even be much lower that of individual samples. Therefore, multi-granularity information processing can perform better robustness than that of a single and finest-granularity. \par

\subsection{Overview of our method}
For image classification tasks, a deep CNN model can be divided into the feature-learning module (FLM) and classifier, and FLM converts the input images into low-dimensional feature vectors, based on which the classifier predicts the label of each individual image sample. 
In this study, we design a GBC module and integrate it into FLM and classifier modules. And we develop an experience replay strategy to train the model, which requires the input images to be divided into empirical and non-empirical samples.
As shown in Fig. \ref{fig3}, (1) through the FLM, two types of input images are converted into a set of low-dimensional feature vectors; (2) each empirical sample from the experience pool is not required to reproduce the $gb$ sample, and the center vector of each empirical $gb$ sample is updated using the feature vectors of individual samples belonging to that $gb$ sample that have just been updated; (3) the GBC module splits the feature vector set of non-empirical samples into the MG grains (i.e., $gb$ samples), each of which contains different number individual samples and corresponds to one single label; (4) a portion of high-purity $gb$ samples is placed as empirical $gb$ samples into the experience pool; and (5) two types of $gb$ samples are merged, and classifier predicts the label of each $gb$ sample rather than the individual sample in the training process. In the error backpropagation of the GBC Layer, we adopt a similar average pooling operation to copy the error of the gb samples to all individual samples within it. In the reasoning process, each individual sample can be considered to be one $gb$ sample. \par

\subsection{Adaptive $gb$ Sample Generation}            
\noindent \textbf{Definition 1:} Given a granular-ball sample $gb_i$, it can contain individual samples with different labels, each of label can correspond different number individual samples. We define $label_{{j}}$ that corresponds the most individual samples as the label of this $gb_i$, $\lvert label_j \rvert$ as the number of individual samples with $label_j$ in $gb_i$, $\lvert gb_i \rvert$ as the number of individual samples in $gb_i$, and $p_gb_i$ as the purity of $gb_i$, then:
\begin{eqnarray}
	p_{{gb_i}}=\frac{\lvert label_{{j}} \rvert}{\lvert gb_{{i}} \rvert} 
	\label{eqPuirty} 
\end{eqnarray}

\textbf{Definition 2:} A set of a low-dimensional feature vector ${D} \in {R}^{{d}}$ is given. We define $C$ as the center of gravity of all individual sample points in a $gb$ sample $gb_i$, $v_{{i}}$ as the feature vector of an individual sample in $gb_i$, and $v_{{c}}$ as the center vector of $gb_i$, then: 

\begin{eqnarray}
	v_{c}=\frac{1}{\lvert gb_{{i}} \rvert} \sum_{i=1}^{\lvert gb_{{i}} \rvert} v_{i}.  
	\label{eqCenter} 
\end{eqnarray}

A formal description of the proposed GBC module is expressed in Eq. \ref{EQ00}. $N$ denotes the total number of samples in the input, $m$ denotes the number of $gb$ sample divided by the input. The construction of $gb$ sample needs to meet the following constraints: (1) each $gb$ sample meets the purity requirements; (2) each $gb$ sample should cover as many samples as possible, and its number should be as few as possible. The purpose of the $GBC$ module is to divide a single-granularity input into a multi-granularity (MG) representation at the feature level. The overall process is summarized in \textbf{Algorithm \ref{alg:algorithm1}.}

\begin{equation}
	\begin{split}
		& f(x, w) \rightarrow g(gb, \theta), \\
		\textbf{s.t.} \ & Min \ N / \sum_{j=1}^{m}(\lvert gb_i \rvert) + m, \\
		& s.t. \ quality(gb_i) \geq T.
	\end{split}
	\label{EQ00}
\end{equation}

\begin{algorithm}[h]
	\caption{Adaptive MG Grain Generation}
	\label{alg:algorithm1}
	\textbf{Input}: feature vector set of input $Z$, purity threshold $p$ \\
	\textbf{Output}: granular-ball set $GBs$, center vector set $CVs$ of granular-ball \\
	\begin{algorithmic}[1]
		\STATE Initializing: $GBs = \emptyset$, $CVs = \emptyset$;
		\STATE $Q.enqueue(Z)$; \textcolor{blue}{\COMMENT{Initialize queue $Q$ with input $Z$ as the first $gb$}}
		\WHILE{$Q$ is not empty}
		\STATE ${gb_i} = Q.dequeue()$; \textcolor{blue}{\COMMENT{Dequeue the first $gb$ from $Q$}} 
        \STATE Compute the purity $p_{{gb_i}}$ of ${gb_i}$, according to Eq. \ref{eqPuirty}
		\IF {$p_{gb_i} \leq p$}
		\STATE \textbf{[$sgb_1$, $sgb_2$]} = 2-means($gb_i$); \textcolor{blue}{\COMMENT{Divide $gb_i$ into two sub-balls using 2-means}} 
		\STATE $Q.enqueue(sgb_1)$;
		\STATE $Q.enqueue(sgb_2)$;
		\ELSE
		\STATE Compute the center vector ${v_c}$ of ${gb_i}$, according to Eq. \ref{eqCenter};
		\STATE $CVs =CVs \cup  v_c$;
		\STATE $GBs =GBs \cup  gb_i$;
		\ENDIF
		\ENDWHILE
		\STATE \textbf{return} $GBs$, $CVs$
	\end{algorithmic}
\end{algorithm}

\subsection{Error Backpropagation in GBC Layer}
The batch input of $N_b$ samples can be mapped to be $N_b$ $d_0$-dimensional feature vectors ([$N_b$, $d_0$]) through the FLM, and GBC layer further divides them into $N_{{gb}}$ $gb$ samples ([$N_{{gb}}$, $d_{{0}}$]), usually $N_{{b}}>N_{{gb}}$. Because of the inconsistency between the input of the FLM and the classifier, error propagation is interrupted between GBC and FLM. Consequently, the error corresponding to each $gb$ sample is returned to the GBC layer during the backpropagation process. However, only the error corresponding to each individual sample ensures that the learning module learns layer-by-layer. The GBC layer performs a similar average pooling operation at the feature level of the input samples. Therefore, we adopt a similar operation to copy the error of the $gb$ samples to all individual samples within it.

\subsection{Experience Replay}
Since the individual samples for each iteration are drawn randomly, $gb$ samples generated for each iteration can exhibit non-static distribution. Consequently, reusing past experience not only reduces training costs but also enables better fitting of the model. Therefore, we design an experience replay strategy that stores the previous $gb$ samples to address these problems. The overall process is summarized as: 

(1) In each training step, we first randomly select a certain number of empirical $gb$ samples from the experience pool and extract the original samples ($X_{{empirical}}$) contained in these empirical $gb$ samples; we also randomly select a certain number of non-empirical samples ($X_{{non-empirical}}$) directly from the training set; batch data [$X_{{empirical}}$, $X_{{non-empirical}}$] are finally fed into the proposed models.

(2) In forward process, it is not necessary to generate $gb$ samples for the empirical samples, and the center vector of each empirical $gb$ sample can be updated using the current feature vector; for non-empirical samples, we generate $gb$ samples through the GBC module and place a portion of high-purity $gb$ samples as empirical $gb$ samples into the experience pool, and the original samples contain in these empirical $gb$ samples are called empirical samples; finally, we merge the two types of $gb$ samples and fed them into the classifier.


\section{Experiments}
We applied our method on base ResNet (RN) \cite{He_Zhang_Ren_Sun_2016}, DenseNet (DN) \cite{huang2017densely} and contrastive learning \cite{Li_Xiong_Hoi_2021,9879400} models, and then we conducted experiments on several image classification datasets (including CIFAR-10, CIFAR-100, CIFAR-10N and ANIMAL-10N). Among them, the noise in CIFAR-10 and CIFAR-100 is generated by random methods, while the noise in CIFAR-10N and ANIMAL-10N is generated by manual annotation. \par

\subsection{Experiments Settings}
\noindent\textbf{Dataset.} For \textbf{CIFAR-10} and \textbf{CIFAR-100} datasets, we test two types of label noise: symmetric noise(Sym.) and asymmetric noise(Asym.). For symmetric noise, a fixed proportion of samples being randomly selected from each category for random label modification; for asymmetric noise, we flipped labels between DEER$\leftrightarrow$HORSE, BIRD$\leftrightarrow$AIRPLANE, TRUCK$\leftrightarrow$AUTOMOBILE, and CAT$\leftrightarrow$DOG(Asym.). 
\textbf{ANIMAL-10N} dataset contains 5 pairs of confusing ANIMAL with atotal of 55,000 images, which are crawled from several online search engines using the predifined labels as the search keyword; the images are then classified by 15 recruited participants; each participant annotated a total of 6,000 images with 600 images per-class; after removing irrelevant images, the training dataset contains 50,000 images and the test dataset contains 5,000 images; the noise rate is about 8\% \cite{song2019selfie}.
\textbf{CIFAR-10N}, variations of CIFAR-10 with human-annotated real-world noisy labels collected from Amazon's Mechanical Turk \cite{wei2022learning}. 

\noindent\textbf{Implementation details.} We implemented the proposed method in PyTorch and conducted experiments on a 24 GB NVIDIA RTX 3090 GPU. We used SGD with Nesterov momentum and set the initial learning rate to 0.1, momentum to 0.9, and minibatch size to 512-1024. The learning rate was dropped by 0.1 at 32k and 48k iterations, and we trained for 64k iterations. The basic models used in the experiments were ResNet and DenseNet models. We used cross-entropy losses with a weight decay of 0.0001. For GBC layer setting, the purity $p$ was set to a value between 0.6 and 1.\par

\noindent\textbf{Baseline methods.} To evaluate our method, we also compared our method to other methods that also without additional data and optimization: (1) CE, which uses Cross-Entropy loss to train the DNNs on noisy datasets. (2) Forward \cite{patrini2017forward}, which corrects loss values by a label transition matrix. (3) LIMIT \cite{pmlr-v119-harutyunyan20a}, which introduces noise into the gradient to avoid memorization. (4) SLN \cite{Chen2021NoiseAN}, which proposes to combat label noise by adding noise to the data labels. (5) CTRR \cite{9879400},  which proposes a contrastive regularization function to learn robust contrastive representations of data over noisy data. \par

\begin{table*}[htbp]
	\centering
	\caption{Comparisons of our GB\_CNN and the original CNN models on CIFAR-10, in which the noise is generated by random method.}
	\renewcommand{\arraystretch}{1.5}
	
	\resizebox{\linewidth}{!}{
		\begin{tabular}{cccccc}
			
			\toprule[1pt]
			
			
			\multirow{2}{*}{\textbf{Models}}	&\multicolumn{4}{c}{\textbf{CIFAR-10}}\\
			\cmidrule(r){2-6} 
			& 0\%  	& 10\%   & 20\%  & 30\%   & 40\%  	\\
			\midrule
			RN20/GB\_RN20	
			& 92.63\begin{tiny}{$\pm$0.02}\end{tiny}/\textbf{92.73\begin{tiny}{$\pm$0.14}\end{tiny}}

			& 90.32\begin{tiny}{$\pm$0.16}\end{tiny}/\textbf{91.79\begin{tiny}{$\pm$0.29}\end{tiny}}

			& 89.06\begin{tiny}{$\pm$0.03}\end{tiny}/\textbf{90.90\begin{tiny}{$\pm$0.12}\end{tiny}}
			
			& 87.30\begin{tiny}{$\pm$0.36}\end{tiny}/\textbf{89.42\begin{tiny}{$\pm$0.11}\end{tiny}}

			& 85.48\begin{tiny}{$\pm$0.08}\end{tiny}/\textbf{88.04\begin{tiny}{$\pm$0.24}\end{tiny}}
			\\

			RN32/GB\_RN32	
			& \textbf{93.63\begin{tiny}{$\pm$0.22}\end{tiny}}/93.50\begin{tiny}{$\pm$0.34}\end{tiny}
			
			& 90.89\begin{tiny}{$\pm$0.23}\end{tiny}/\textbf{92.35\begin{tiny}{$\pm$0.05}\end{tiny}}

			& 89.67\begin{tiny}{$\pm$0.21}\end{tiny}/\textbf{91.20\begin{tiny}{$\pm$0.30}\end{tiny}}
			
			& 87.76\begin{tiny}{$\pm$0.18}\end{tiny}/\textbf{90.09\begin{tiny}{$\pm$0.28}\end{tiny}}
			
			& 85.71\begin{tiny}{$\pm$0.05}\end{tiny}/\textbf{88.21\begin{tiny}{$\pm$0.04}\end{tiny}}
			
			\\
			
			RN44/GB\_RN44	
			& \textbf{93.78\begin{tiny}{$\pm$0.08}\end{tiny}}/93.73\begin{tiny}{$\pm$0.25}\end{tiny}	
			
			& 91.05\begin{tiny}{$\pm$0.17}\end{tiny}/\textbf{92.49\begin{tiny}{$\pm$0.18}\end{tiny}}

			& 89.43\begin{tiny}{$\pm$0.11}\end{tiny}/\textbf{91.43\begin{tiny}{$\pm$0.10}\end{tiny}}
			
			& 87.91\begin{tiny}{$\pm$0.44}\end{tiny}/\textbf{89.89\begin{tiny}{$\pm$0.16}\end{tiny}}

			& 85.99\begin{tiny}{$\pm$0.14}\end{tiny}/\textbf{88.02\begin{tiny}{$\pm$0.23}\end{tiny}}
			
			\\
			
			RN56/GB\_RN56
			& 94.05\begin{tiny}{$\pm$0.37}\end{tiny}/\textbf{94.23\begin{tiny}{$\pm$0.18}\end{tiny}}	
			
			& 90.83\begin{tiny}{$\pm$0.16}\end{tiny}/\textbf{92.40\begin{tiny}{$\pm$0.44}\end{tiny}}

			& 89.75\begin{tiny}{$\pm$0.15}\end{tiny}/\textbf{91.49\begin{tiny}{$\pm$0.46}\end{tiny}}
			
			& 87.76\begin{tiny}{$\pm$0.47}\end{tiny}/\textbf{90.08\begin{tiny}{$\pm$0.10}\end{tiny}}

			& 85.96\begin{tiny}{$\pm$0.09}\end{tiny}/\textbf{88.22\begin{tiny}{$\pm$0.25}\end{tiny}}
			
			\\
			
			DN121/GB\_DN121	
			& \textbf{95.39\begin{tiny}{$\pm$0.04}\end{tiny}}/94.04\begin{tiny}{$\pm$0.13}\end{tiny}	
			
			& 90.38\begin{tiny}{$\pm$0.21}\end{tiny}/\textbf{92.64\begin{tiny}{$\pm$0.07}\end{tiny}}
			
			& 85.28\begin{tiny}{$\pm$0.56}\end{tiny}/\textbf{90.77\begin{tiny}{$\pm$0.29}\end{tiny}}
			
			& 82.63\begin{tiny}{$\pm$0.29}\end{tiny}/\textbf{88.61\begin{tiny}{$\pm$0.21}\end{tiny}}

			& 80.90\begin{tiny}{$\pm$0.26}\end{tiny}/\textbf{85.52\begin{tiny}{$\pm$0.33}\end{tiny}}
			
			\\

			\bottomrule[1pt]
		\end{tabular}%
	}	
	\label{tab:tab1}%
\end{table*}%

\subsection{Comparisons with the Original Models}
\begin{figure}[h]
	\centering
	\includegraphics[width=0.85\textwidth]{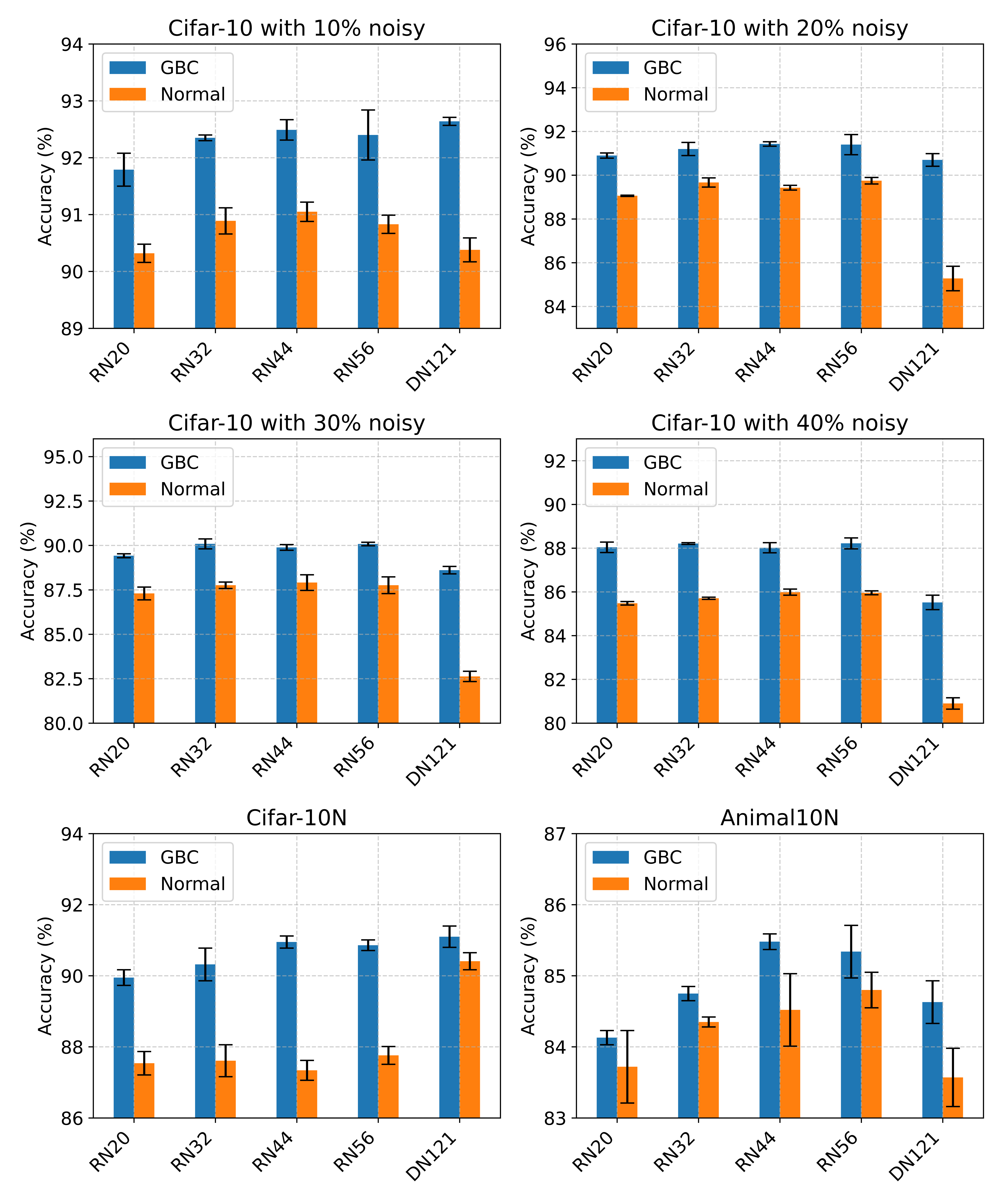}
	\caption{Comparisons of our GB\_CNN and the original CNN models.} 
	\label{fig2}
\end{figure}

\noindent We first applied the proposed GBC module on two classical models (ResNet and DenseNet) —and trained them on the benchmark dataset which contained different proportions of labeled noise. The comparison results of the original CNN and our GB\_CNN models are listed in Table \ref{tab:tab1}, Table \ref{tab:tab4} and Fig. \ref{fig2}. From the results, we can make two major observations:

\begin{table}[htbp]
	\centering
	\caption{Comparisons of our GB\_CNN and the original CNN models on CIFAR-10N and ANIMAL-10N, in which the noise is generated by manual annotation.}
	\renewcommand{\arraystretch}{1.2}
	
	\resizebox{0.7\linewidth}{!}{
		\begin{tabular}{ccc}
			
			\toprule[1pt]
			
			\multirow{2}{*}{\textbf{Models}}	&\multicolumn{1}{c}{\textbf{CIFAR-10N}}	& \multicolumn{1}{c}{\textbf{ANIMAL-10N}}\\
			\cmidrule(r){2-2} \cmidrule(r){3-3}
			
			& 9.03\%    	& 8\%\\
			
			\midrule
			RN20/GB\_RN20	
			& 87.54\begin{tiny}{$\pm$0.33}\end{tiny}/\textbf{89.95\begin{tiny}{$\pm$0.22}\end{tiny}}
			& 83.72\begin{tiny}{$\pm$0.51}\end{tiny}/\textbf{84.13\begin{tiny}{$\pm$0.10}\end{tiny}}
			\\
			
			RN32/GB\_RN32	
			& 87.61\begin{tiny}{$\pm$0.45}\end{tiny}/\textbf{90.32\begin{tiny}{$\pm$0.46}\end{tiny}}
			& 84.35\begin{tiny}{$\pm$0.07}\end{tiny}/\textbf{84.75\begin{tiny}{$\pm$0.10}\end{tiny}}
			\\
			
			RN44/GB\_RN44	
			& 87.34\begin{tiny}{$\pm$0.28}\end{tiny}/\textbf{90.95\begin{tiny}{$\pm$0.17}\end{tiny}}
			& 84.52\begin{tiny}{$\pm$0.51}\end{tiny}/\textbf{85.48\begin{tiny}{$\pm$0.11}\end{tiny}}
			\\
			
			RN56/GB\_RN56
			& 87.76\begin{tiny}{$\pm$0.25}\end{tiny}/\textbf{90.86\begin{tiny}{$\pm$0.15}\end{tiny}}
			& 84.80\begin{tiny}{$\pm$0.25}\end{tiny}/\textbf{85.34\begin{tiny}{$\pm$0.37}\end{tiny}}
			\\
			
			DN121/GB\_DN121	
			& 90.41\begin{tiny}{$\pm$0.24}\end{tiny}/\textbf{91.10\begin{tiny}{$\pm$0.30}\end{tiny}}
			& 83.57\begin{tiny}{$\pm$0.41}\end{tiny}/\textbf{84.63\begin{tiny}{$\pm$0.30}\end{tiny}}
			\\
			
			\bottomrule[1pt]
		\end{tabular}%
	}	
	\label{tab:tab4}%
\end{table}%

(1) The purpose of our proposed method is not to push the state-of-the-art performance of the original models leanring on clean data, but to reduce the influence of label noise. From the results in Table \ref{tab:tab1}, we can note that the CNN models with embedding the GBC module can perform almost as well as the original models in terms of learning with no label noise, that is, the proposed GBC layer does not decrease the performance of the original models. Because, our method does not filter or penalize mislabeled samples, and all samples are used for learning.

\begin{table}[htbp]
	\centering
	\caption{The average proportion of label noise before and after the GBC module in training with CIFAR-10.}
	\renewcommand{\arraystretch}{1.2}
	\tabcolsep=0.35cm
	\resizebox{0.7\linewidth}{!}{
		
		\begin{tabular}{cccccc}
			\toprule[0.9pt]
			\multirow{1}{*}[-2ex]{\textbf{Models}}	& \multicolumn{5}{c}{\textbf{Label Noise Rate(\%) of Training Dataset}}\\
			\cmidrule{2-6} & 10     & 20    & 30    & 40    & 50  \\
			\midrule
			GB\_RN20	&  1.32   &  4.26    &  7.23    	& 14.23    & 21.60  \\
			GB\_RN32	&  1.10   &  3.94    &  7.68    	& 13.37    & 20.61  \\
			GB\_RN44	&  1.44   &  3.99    &  7.87    	& 13.62    & 20.79  \\
			GB\_RN56	&  1.38   &  3.96    &  8.01    	& 13.82    & 21.06  \\
			GB\_DN121 	&  1.69   &  5.29    & 10.17        & 15.60    & 24.09  \\
			\bottomrule[0.9pt]
		\end{tabular}%
	}
	\label{tab:tab2}%
\end{table}%

(2) On learning with varying proportion of labeled noise (See Table \ref{tab:tab1}), our method can significantly the robustness of the original deep model with no additional data and optimization. One fundamental reason for this is that the label-noise proportion of the $gb$ samples in the training process is considerably lower than that of the individual samples, as shown in Table \ref{tab:tab2}. We can conclude that our method improves the robustness of CNN models by reducing the noise ratio of training samples during the training process, without any additional data and optimization. In addition, our method can also improve the robustness of the models on datasets that generate noise through manual annotation, which conforms to the natural distribution of noise (See Table. \ref{tab:tab4}).\par

\begin{table}[htbp]
	\centering
	\caption{Several CNN models used in the experiments.}
	
	\renewcommand{\arraystretch}{1.1}
	
	\resizebox{0.5\linewidth}{!}{
		
		\begin{tabular}{ccccccc}
			\toprule[1pt]
			\textbf{Models} 			& RN20	& RN32 & RN44 & RN56 & RN18 & DN121\\
			\midrule
			\textbf{params(M)}		& 0.27   & 0.46    & 0.66    & 0.85    & 11.69		& 7.98\\
			\bottomrule[1pt]
		\end{tabular}%
	}
	\label{tab:tab0}%
\end{table}%

\subsection{Comparisons with Other Methods}

Since, PreAct ResNet-18 (PRN18, see Table \ref{tab:tab0}), a much wider and larger model compared with RN20, 32, 44, 56 and DN121, was used to construct the experiments in the previous studies, and thus, to ensure fairness in comparison, we also applied our method on the PRN18. In addition, CTRR method achieved the SOTA result in robustness on the label noise learning; therefore, we also applied the proposed method on CTRR framework, in which the GBC module was embedded into the supervised learning branch with no other changes. From the results in Table \ref{tab:tab3}, we can make the major observations as: (1) Our method significantly improves the robustness of the original models (CE and GB\_PRN18), and also perform better on varying noise ratio and different noise type comparing with the listed methods. (2) When the proposed GBC module is embeded into CTRR framework, the new method further improves the robustness of CTRR method and achieves the state-of-the-art results.

\begin{table*}[htbp]
	\centering
	\caption{Comparisons with other methods on CIFAR-10.}
	\renewcommand{\arraystretch}{1.3}
	\resizebox{\linewidth}{!}{
		
		\begin{tabular}{ccccccccc}	
			\toprule[1pt]			
			\multirow{3}{*}[-1.5ex]{\textbf{Models(Arch)}}	&\multicolumn{4}{c}{\textbf{CIFAR-10}} & \multicolumn{3}{c}{\textbf{CIFAR-100}}	& \multicolumn{1}{c}{\textbf{Animal-10N}}\\
			\cmidrule(r){2-5}	\cmidrule(r){6-8} \cmidrule(r){9-9}
			& \multicolumn{3}{c}{\textbf{Sym.}}	&	\multicolumn{1}{c}{\textbf{Asym.}} &	\multicolumn{3}{c}{\textbf{Sym.}} & \multicolumn{1}{c}{\multirow{2}{*}{8\%}}\\
			\cmidrule(r){2-4} \cmidrule(r){5-5} \cmidrule(r){6-8}
			
			& 0\%   & 20\%  & 40\%  & 40\%  & 0\%   & 20\%  & 40\% &	\\
			
			\midrule
			CE(PRN18)
			& 93.97\small{$\pm$0.22} 	& 88.51\small{$\pm$0.17} 	& 82.73\small{$\pm$0.16}		& 83.23\small{$\pm$0.59}
			& 73.21\small{$\pm$0.14} 	& 60.57\small{$\pm$0.53} 	& 52.48\small{$\pm$0.34}		
			& 83.18\small{$\pm$0.15}
			\\
			
			Forward \cite{patrini2017forward}(PRN18)
			& 93.47\small{$\pm$0.19} 	& 88.87\small{$\pm$0.21} 	& 83.28\small{$\pm$0.37}		& 82.93\small{$\pm$0.74}	
			& 73.01\small{$\pm$0.33} 	& 58.72\small{$\pm$0.54} 	& 50.10\small{$\pm$0.84}
			& 83.67\small{$\pm$0.31}
			\\
			
			
			LIMIT \cite{pmlr-v119-harutyunyan20a}(PRN18)
			& 93.47\small{$\pm$0.56} 	& 89.63\small{$\pm$0.42} 	& 85.39\small{$\pm$0.63}		& 83.56\small{$\pm$0.70}
			& 65.53\small{$\pm$0.91} 	& 58.02\small{$\pm$1.93} 	& 49.71\small{$\pm$1.81}
			& -	
			\\
			
			SLN \cite{Chen2021NoiseAN}(PRN18)
			& 93.21\small{$\pm$0.21} 	& 88.77\small{$\pm$0.23} 	& 87.03\small{$\pm$0.70}		& 81.02\small{$\pm$0.25}
			& 63.13\small{$\pm$0.21} 	& 55.35\small{$\pm$1.26} 	& 51.39\small{$\pm$0.58}
			
			& 83.17\small{$\pm$0.08}	
			\\
			
			A-PolySoft  \cite{shu2020learning}(RN32)
			& 92.12\small{$\pm$0.12} 	& 89.73\small{$\pm$0.20} 	& 87.22\small{$\pm$0.36}		& -
			& - & - & - 
			& - 
			\\
			
			GB\_PRN18(ours)
			& 94.25\small{$\pm$0.56}	& 91.87\small{$\pm$0.24}	& 88.56\small{$\pm$0.29}	& 83.92\small{$\pm$0.87}
			& 74.14\small{$\pm$0.50}	& 63.66\small{$\pm$0.39}	& 59.22\small{$\pm$1.44}
			& 85.51\small{$\pm$0.22} 
			\\
			
			\midrule
			CTRR \cite{9879400}(PRN18)
			& 94.29\small{$\pm$0.21}	& 93.05\small{$\pm$0.32}	& 92.16\small{$\pm$0.31}		& 89.00\small{$\pm$0.56}
			& 74.36\small{$\pm$0.41}	& 70.09\small{$\pm$0.45}	& 65.32\small{$\pm$0.20} 
			& 86.71\small{$\pm$0.15}
			\\
			GB\_CTRR(ours)		
			& \textbf{95.13\small{$\pm$0.22}}	& \textbf{94.24\small{$\pm$0.28}}	& \textbf{93.20\small{$\pm$0.09}}	& \textbf{89.87\small{$\pm$0.30}}
			& \textbf{75.75\small{$\pm$0.48}}	& \textbf{70.21\small{$\pm$0.32}}	& \textbf{66.92\small{$\pm$0.26} }
			& \textbf{87.05\small{$\pm$0.27}}
			\\
			
			\bottomrule[1pt]
		\end{tabular}%
	}	
	\label{tab:tab3}%
\end{table*}%

\section{Conclusion}
In the practice, a certain proportion of samples with wrong labels always occurs when collecting data, which can affect the effectiveness of models. Consequently, labels can often change due to subjective factors, while the content of the sample or its feature do not change with changes in the labeling. Inspired by this, we propose learning the multi-granularity representations based on the feature similarity, where the classifier can predict the label of each $gb$ sample instead of the individual samples. The experimental results verify that the proposed method can improve the robustness of deep CNN models without any additional data and optimization. Nevertheless, our proposed still needs improvement in classification tasks with many categories, which is worth further exploration.

\bibliographystyle{splncs04}
\bibliography{mybibfile}

\end{document}